# LASSO-MOGAT: A Multi-Omics Graph Attention Framework for Cancer Classification

*Fadi Alharbi [1], Aleksandar Vakanski [1,*], Murtada K. Elbashir [2] and Mohanad Mohammed [3]*

Affiliation: [1]College of Engineering, Department of Computer Science, University of Idaho, Moscow, ID 83844, USA; alha5622@vandals.uidaho.edu, vakanski@uidaho.com

[2]College of Computer and Information Sciences, Department of Information Systems, Jouf University, Sakaka, Aljouf 72441, Saudi Arabia; mkelfaki@ju.edu.sa

[3]School of Mathematics, Statistics and Computer Science, University of KwaZulu-Natal, Pietermaritzburg, Scottsville, 3209, South Africa; mohanadadam32@gmail.com

* Correspondence: vakanski@uidaho.com

**Abstract**: The application of machine learning methods to analyze changes in gene expression patterns has recently emerged as a powerful approach in cancer research, enhancing our understanding of the molecular mechanisms underpinning cancer development and progression. Combining gene expression data with other types of omics data has been reported by numerous works to improve cancer classification outcomes. Despite these advances, effectively integrating high-dimensional multi-omics data and capturing the complex relationships across different biological layers remains challenging. This paper introduces LASSO-MOGAT (LASSO-Multi-Omics Gated ATtention), a novel graph-based deep learning framework that integrates messenger RNA, microRNA, and DNA methylation data to classify 31 cancer types. Utilizing differential expression analysis with LIMMA and LASSO regression for feature selection, and leveraging Graph Attention Networks (GATs) to incorporate protein-protein interaction (PPI) networks, LASSO-MOGAT effectively captures intricate relationships within multi-omics data. Experimental validation using five-fold cross-validation demonstrates the method's precision, reliability, and capacity for providing comprehensive insights into cancer molecular mechanisms. The computation of attention coefficients for the edges in the graph by the proposed graph-attention architecture based on protein-protein interactions proved beneficial for identifying synergies in multi-omics data for cancer classification.

## 1. Introduction

Recent advances in machine learning (ML) and deep learning (DL) approaches have advanced our capacity to categorize various forms of cancer [1]. Graph-based DL architectures particularly offer the potential to leverage complex biological networks, such as protein-protein interaction (PPI) networks and gene regulatory networks, for extracting meaningful data representations. By treating biological networks as graphs where nodes represent biological entities (e.g., genes, proteins) and edges represent interactions between them, graph-based architectures effectively capture the topological and functional characteristics of biological networks [2]. Graph-based architectures based on graph neural networks (GNNs), graph convolutional neural networks (GCNNs), graph attention networks (GATs), and graph transformer networks (GTNs) have been successfully utilized for integrating multi-omics data and biological networks [3,4], and have shown promise for learning complex patterns from biological networks and omics data [5]. For instance, GATs employ attention mechanisms to focus on important nodes and edges in a graph, providing insights into the molecular mechanisms underlying cancer development and progression [6]. The utilization of graph-based networks in multi-omics data analysis is likely to continue to grow, with further advancements in model architectures and applications [7].

Gene expression data is a valuable resource in cancer research, offering insights into the activity levels of genes in specific tissues or cell types, and enabling comparisons between cancerous and normal cells. [8]. By measuring the amount of messenger RNA (mRNA) produced by each gene, researchers can determine which genes are being actively transcribed and expressed in these specific contexts [9]. This information is crucial for understanding the molecular mechanisms underlying cancer development and progression, where certain genes may be upregulated (increased expression) or downregulated (decreased expression) compared to normal cells [10]. The changes in gene expression can be indicative of the dysregulation of cellular processes, such as cell growth, division, and apoptosis, which are distinguishing features of cancer. By analyzing gene expression patterns, researchers can identify genes that are specifically altered in different cancer types, providing valuable biomarkers for early detection and diagnosis.

To gain a more comprehensive understanding of cancer biology, researchers have integrated gene expression data with other types of omics data, such as DNA methylation and microRNA (miRNA) expression data [11]. The multi-omics approach allows to explore the complex relationships between different molecular layers and identify key molecular alterations that drive cancer development and progression [12]. For example, DNA methylation data can reveal epigenetic changes that silence tumor suppressor genes or activate oncogenes, contributing to cancer initiation and advancement [13]. MiRNA expression data, on the other hand, can offer insights into post-transcriptional gene regulation and its role in cancer pathogenesis [14]. Overall, the integration of gene expression with multi-omics data offers a more inclusive view of cancer biology and holds great promise for advancing precision oncology.

This paper introduces LASSO-MOGAT, a graph-based DL approach for integrating multi-omics data comprising mRNA, miRNA, and DNA methylation data for classification of 31 cancer types. To select the most informative multi-omics features, we implemented data preprocessing using differential expression analysis (DEG) with LIMMA (Linear Models for Microarray) [15], which involves fitting a linear model to each gene to estimate the difference in expression between conditions while accounting for variability. LIMMA uses an empirical Bayes approach to moderate the standard errors of the estimated log-fold changes, improving the accuracy of differential expression detection, particularly for genes with low expression levels. This method provides robust statistical inference for identifying genes that are differentially expressed across conditions with high sensitivity and specificity. We applied LASSO regression to further refine the feature selection process by penalizing the absolute size of the regression coefficients, encouraging sparsity, and selecting the most relevant features for the classification task. We used the extracted lower-dimensional data representation as inputs to a Graph Attention Network (GAT), leveraging the topological information from protein-protein interaction networks for classification. Experimental validation of the proposed approach using five-fold cross-validation demonstrates the efficacy of the introduced integration of multi-omics data and graph-based GAT model for precise and reliable cancer classification.

Unlike other studies for cancer classification that focus on single types of omics data or simpler feature selection methods [16,17], the proposed LASSO-MOGAT approach employs differential expression analysis with LIMMA and LASSO regression for robust feature selection. The employed GATs model effectively captures complex relationships within multi-omics data by incorporating PPI networks. Additionally, our approach differs from other studies based on GATs, such as MODIG [18], which aims to identify cancer driver genes by integrating multi-omics pan-cancer data, including mutations and copy number variants, with multi-dimensional gene networks. MODIG leverages GATs to generate gene representations across different dimensions, including PPI, gene sequence similarity, and KEGG pathway co-occurrence, to enhance the identification of driver genes. While both approaches utilize GATs for processing multi-omics data, their objectives differ since the focus of our work is on cancer type classification, whereas MODIG [18] is designed for cancer driver gene identification.

Our approach also stands apart from methods such as GOAT [19], MORGAT [20], DGP-AMIO [21], and MOGAT [7], which either target different diseases (e.g., asthma) or employ distinct graph-based learning strategies. Furthermore, LASSO-MOGAT is distinct from works that utilize GNNs for processing multi-omics data but focus on different aspects and methodologies. For instance, Li et al. [6] introduced a framework that combines both inter-omics and intra-omics connections using a heterogeneous multi-layer graph, emphasizing the integration of different types of GNNs (GCN and GAT) with the focus on classification of cancer subtypes.

The key contributions of this work to the field of cancer classification are as follows:
- We propose a method to integrate RNA-Seq, miRNA, and DNA methylation data for 31 cancer types in addition to normal samples, providing a comprehensive multi-omics perspective on cancer.
- We develop a GAT model that leverages protein-protein interaction (PPI) networks to classify cancer types, including normal samples, effectively capturing the complex relationships within the multi-omics data.
- We empirically demonstrate that the use of DEGs with LIMMA and LASSO regression is effective in selecting the most relevant features from multi-omics data, significantly improving the performance of the classification models.

## 2. Related Works

In recent years, the integration of multi-omics data has emerged as a promising approach in cancer research, offering deeper insights into the complex biological mechanisms underlying the disease. Several studies have proposed novel algorithms and models to effectively integrate and analyze multi-omics data for cancer subtype classification, leveraging deep learning (DL) techniques and attention mechanisms to enhance model interpretability and performance.

Mostavi et al. [22] introduced several Convolutional Neural Network (CNN) models for cancer type prediction based on gene expression data. Their study presented three different CNN architectures—1D-CNN, 2D-Vanilla-CNN, and 2D-Hybrid-CNN—that demonstrated the importance of addressing tissue origin effects in cancer marker identification. This work highlighted that CNN models could effectively classify cancer types and identify cancer marker genes through model interpretation techniques, making significant strides in cancer type prediction using gene expression profiles.

Ramirez et al. [23] applied Graph Convolutional Networks (GCNs) for classifying cancer types using gene expression data. They utilized different graph structures, including co-expression and protein-protein interaction (PPI) graphs, to capture molecular interactions and identify cancer-specific marker genes. Their work demonstrated the power of GCNs in leveraging multi-omics data for cancer classification and marker identification, achieving high prediction accuracies and providing insights into the effects of gene perturbations.

Kaczmarek et al. [24] developed a Graph Transformer Network (GTN) for cancer classification, leveraging miRNA and mRNA expression data to model biological interactions and targeting pathways. The GTN provided high interpretability through self-attention mechanisms, identifying important pathways and biomarkers. While their GTN did not outperform all baseline models in terms of accuracy, its interpretability offers valuable insights into miRNA-mRNA interactions, making it a crucial tool for understanding complex biological relationships in cancer.

Other significant contributions include Moon et al. [25], who proposed the MOMA (Multi-Omics Module Analysis) model for cancer subtype classification using a geometrical approach and attention mechanisms to integrate multi-biomedical data. Zhang et al. [26] presented a DL method integrating multi-omics data using similarity network fusion and a graph autoencoder, enhancing performance on TCGA datasets. Sun et al. [27] introduced SADLN (Self-Attention based Deep Learning Network) for cancer subtype recognition, integrating multi-omics data and modeling sample relationships based on latent low-dimensional representations.

Shanthamallu et al. [28] presented GrAMME (Graph Attention Models for Multi-layered Embeddings) for semi-supervised learning with multi-layered graphs, developing two architectures to leverage inter-layer dependencies. Qiu et al. [16] introduced GGAT (Gated Graph Attention Network) for cancer prediction, combining gating and attention mechanisms to improve correlation capture and enhance prediction accuracy. Zhao et al. [18] proposed MODIG, integrating multi-omics data with multi-dimensional gene networks using GATs for effective cancer research. Baul et al. [17] developed omicsGAT for RNA-seq data in cancer subtyping, using graph-based learning and attention mechanisms for better subtype prediction. Ouyang et al. [29] proposed MOGLAM, integrating multi-omics data for disease classification and biomarker identification using dynamic graph convolution and attention mechanisms. Gong et al. [30] introduced MOADLN, using self-attention to capture patient correlations within omics types and cross-omics correlations. Song et al. [31] presented GGraphSAGE, integrating multi-omics data with GNNs to predict cancer driver genes, demonstrating effectiveness across multiple tumor types.

Autoencoders have been widely used for dimensionality reduction and feature extraction in multi-omics data analysis. Zhang et al. [32] utilized a variational autoencoder for classifying different cancer types using RNA-seq gene expression and DNA methylation data. Chai et al. [33] designed DCAP (Denoising Autoencoder for Accurate Cancer Prognosis Prediction) to integrate multi-omics data for cancer prognosis prediction. Li et al. [34] proposed MoGCN (Multi-Omics Graph Convolutional Network), using an autoencoder for feature extraction and similarity network fusion to construct patient similarity networks. Zhou et al. [35] and Khadirnaikar et al. [36] used autoencoders to analyze multi-omics data, reducing dimensionality and identifying novel cancer subtypes through clustering and machine learning models.

Graph Neural Networks (GNNs) have emerged as powerful models for analyzing multi-omics data, effectively capturing the relationships between different biological entities. Zhu et al. [37] proposed GGNN (Geometric Graph Neural Network) incorporating geometric features for improved predictive power and interpretability. Xiao et al. [38] introduced MPKGNN (Multi-Prior Knowledge Graph Neural Network) for multi-omics data analysis, leveraging multiple prior knowledge graphs for cancer molecular subtype classification. Chatzianastasis et al. [39] developed EMGNN (Explainable Multilayer Graph Neural Network) to identify cancer genes by leveraging multiple gene-gene interaction networks and pan-cancer multi-omics data.

In addition to GNNs, various models integrating Transformer and Graph Convolutional Networks (GCNs) have been developed to enhance disease classification accuracy using multi-omics data. Wang et al. [40] introduced MOSEGCN, combining Transformer and Similarity Network Fusion (SNF) for precise disease subtype classification. Yao et al. [41] developed GCNFORMER, a model integrating GCN and Transformer to predict lncRNA-disease associations, demonstrating the effectiveness of combining these approaches for improved performance and interpretability.

Convolutional Neural Networks (CNNs) are primarily utilized for grid-like data, such as images, where the spatial relationships between pixels are crucial for feature extraction. The interpretability of CNNs often involves visualizing filters and understanding which regions of an image activate specific neurons. In contrast, Graph Attention Networks (GATs) are designed for graph-structured data, where nodes represent entities and edges represent relationships, which can vary in connectivity and significance. GATs enhance interpretability through their attention mechanisms, which dynamically weigh the importance of neighboring nodes during message passing. This attention mechanism provides insights into which nodes influence predictions and how relationships are weighted within the graph. Consequently, GATs are particularly adept at interpreting relational data and complex dependencies within graphs. While CNNs emphasize spatial hierarchies in images, GATs focus on relational structures in graphs, thereby offering a different perspective on interpretability.

In summary, the integration of multi-omics data in cancer research has been greatly enhanced by advanced techniques such as autoencoders, GNNs, GCNs, and GATs. These methods have shown remarkable performance in cancer subtype classification, feature extraction, and understanding complex biological mechanisms underlying the disease. They offer improved interpretability, predictive power, and performance compared to traditional methods, making them valuable tools for researchers. Despite these advancements, several challenges remain. Multi-omics data are generated from various platforms, each with unique characteristics and measurement techniques, making their integration complex. The vast number of features compared to the number of samples in these datasets (large p small n issues) can lead to overfitting and reduced model performance, necessitating effective dimensionality reduction techniques. Additionally, multi-omics data often contain noise and sparse measurements, obscuring true biological signals and requiring robust preprocessing methods. Batch effects, or variations between different batches of data, must be corrected to avoid misleading results. While advanced models like GNNs and GATs offer improved performance, their complexity can make interpretation difficult, posing a challenge in understanding the underlying biological mechanisms. Finally, the computational complexity of integrating and analyzing multi-omics data requires substantial resources and efficient algorithms to handle large-scale datasets. Therefore, there is a pressing need to develop new methods that can effectively address these challenges, improve data integration, and enhance our understanding of the molecular mechanisms driving cancer progression.

## 3. Materials and Methods

### 3.1 Data Collection

We retrieved the omics data for the different cancer types from the Pan-Cancer Atlas [42] using the Genomic Data Commons (GDC) query function from the TCGAbiolinks library [43]. GDC is a project by the National Cancer Institute (NCI) that provides unified data storage for cancer genomic studies, facilitating data sharing within the research community.

GDCquery function requires to specify parameters related to project, legacy, data.category, data.type, and sample.type. The project parameter refers to the name of the cancer research project within TCGA from which to retrieve data. TCGA consists of multiple cancer projects, each focused on a specific cancer type or cancer-related research area. In our case, we set the project parameter to "TCGA-*" to specify the 33 TCGA projects which include the normal tissues. The legacy argument was configured as True, signifying that the query was directed to the legacy repository to retrieve the original, unaltered data that had been stored in the TCGA Data Portal. The data.category parameter specifies the specific data category relevant to the project. For our case, we set the data category to transcriptome profiling for the mRNA or RNA-Seq and miRNA datasets, while for the methylation data we specified the data category as DNA Methylation. The data.type parameter defines the specific data type used for filtering the files to be downloaded. We specified this parameter as Gene expression quantification, miRNA Expression Quantification, and Methylation Beta Value for mRNA or RNA-Seq, miRNA, and DNA Methylation data, respectively. The sample.type parameter defines the type of sample used for filtering the data to be downloaded. In our case, we specified the sample type as "c ("Primary Solid Tumor", "Solid Tissue Normal")", indicating that we aimed to download gene expression data specifically from normal cases and cases with tumors. The data was saved into a format where the columns represent the samples or cases, and the rows represent the features of interest for each data type. A summary of the collected data is presented in Table 1.

**Table 1.** Tumor types and number of samples of TCGA Multi-omics data (mRNA, miRNA, and DNA methylation) used in the analysis.

| Available Cancer Types | | Number of Samples | | |
|---|---|---|---|---|
| | | mRNA | miRNA | DNA methylation |
| Uterine corpus endometrial carcinoma | UCEC | 588 | 578 | 484 |
| Adrenocortical carcinoma | ACC | 79 | 80 | 80 |
| Brain lower grade glioma | LGG | 516 | 512 | 516 |
| Sarcoma | SARC | 261 | 259 | 265 |
| Pancreatic adenocarcinoma | PAAD | 182 | 182 | 194 |
| Esophageal carcinoma | ESCA | 197 | 199 | 201 |
| Prostate adenocarcinoma | PRAD | 553 | 550 | 552 |
| Acute Myeloid Leukemia | LAML | - | - | - |
| Kidney renal clear cell carcinoma | KIRC | 613 | 615 | 484 |
| Pheochromocytoma and Paraganglioma | PCPG | 182 | 182 | 182 |
| Head and Neck squamous cell carcinoma | HNSC | 564 | 567 | 578 |
| Ovarian serous cystadenocarcinoma | OV | 421 | 490 | 10 |
| Glioblastoma multiforme | GBM | 163 | 5 | 142 |
| Uterine carcinosarcoma | UCS | 57 | 57 | 57 |
| Mesothelioma | MESO | 87 | 87 | 87 |
| Testicular germ cell tumors | TGCT | 150 | 150 | 150 |
| Kidney chromophobe | KICH | 91 | 91 | 66 |
| Rectum adenocarcinoma | READ | 176 | 164 | 105 |
| Uveal melanoma | UVM | 80 | 80 | 80 |
| Thyroid carcinoma | THCA | 564 | 565 | 563 |
| Liver hepatocellular carcinoma | LIHC | 421 | 422 | 427 |
| Thymoma | THYM | 122 | 126 | 126 |
| Cholangiocarcinoma | CHOL | 44 | 45 | 45 |
| Lymphoid neoplasm diffuses large B-cell lymphoma | DLBC | 48 | 47 | 48 |
| Kidney renal papillary cell carcinoma | KIRP | 322 | 325 | 320 |
| Bladder urothelial carcinoma | BLCA | 431 | 436 | 439 |
| Skin cutaneous melanoma | SKCM | 104 | 99 | 106 |
| Lung squamous cell carcinoma | LUSC | 553 | 523 | 412 |
| Stomach adenocarcinoma | STAD | 448 | 491 | 397 |
| Lung adenocarcinoma | LUAD | 598 | 565 | 505 |
| Colon adenocarcinoma | COAD | 522 | 463 | 350 |
| Cervical squamous cell carcinoma and endocervical adenocarcinoma | CESC | 307 | 310 | 310 |
| Breast invasive carcinoma | BRCA | 1,224 | 1,200 | 890 |
| **Total** | | **10,668** | **10,465** | **9,171** |

## 3.2 Data Preprocessing

### 3.2.1 Differential Gene Expression Analysis

In genomics, differential gene expression analysis is a technique used to identify genes that exhibit different expression levels under two or more biological conditions, such as treatment versus control samples or tumor versus healthy tissue [44]. The goal is to understand how gene expression patterns change in response to various circumstances, providing insights into the underlying biological processes. We conducted differential gene expression analysis for mRNA or RNA-Seq data using the DESeq2 package in R, which fits a generalized linear model to the count data for each gene, assuming a negative binomial distribution to account for biological variability and overdispersion. The Wald test is used to evaluate the significance of the estimated log fold changes, and p-values derived from the Wald statistic are calculated to determine differential expression. By specifying a p-value threshold of 0.001, we focused on identifying statistically significant genes that are likely to play a role in the biological processes under study.

### 3.2.2 LIMMA Model

We employed LIMMA for fitting a linear model to the DNA methylation data, where the methylation levels of CpG sites are modelled as a function of the experimental sample groups [45]. The DNA methylation is from Human Methylation 450K (HM450) [46] and has 485,577 features and 9,171 samples. We applied LIMMA on the DNA methylation data to identify differentially methylated CpG sites between Tumor and Normal samples. LIMMA estimates the effect size (difference in methylation levels between groups) and calculates a moderated t-statistic for each CpG site. The p-value associated with the t-statistics indicates the statistical significance of the differences in methylation levels between groups. In our case, we set the p-value for LIMMA to 0.05 which reduced the number of features in the DNA methylation data to 139,321 features.

### 3.2.3 LASSO Regression Model

LASSO (Least Absolute Shrinkage and Selection Operator) regression is a type of linear regression that uses L1 regularization [47], given with:

$$\text{minimize} \left( \sum_{i=1}^{n}(Y_i - \beta_0 - \sum_{j=1}^{p}\beta_j\, x_{ij})^2 + \lambda \sum_{j=1}^{p}|\beta_j| \right) \quad (1)$$

where $n$ is the number of samples, $p$ is the number of features, $Y_i$ is the target variable for the $i^{th}$ sample, $x_{ij}$ is the $j^{th}$ feature of the $i^{th}$ sample, $\beta_0$ is the intercept, $\beta_j$ is the coefficient for the $j^{th}$ feature, and $\lambda$ is the regularization parameter that controls the strength of the penalty term. The term $\lambda \sum_{j=1}^{p}|\beta_j|$ is the L1 penalty term that encourages sparsity in the coefficient vector, effectively selecting only a subset of the most important features while setting the coefficients of less important features to zero.

We used LASSO regression for feature selection and regularization for both mRNA or RNA-Seq and DNA methylation data. The mRNA or RNA-Seq data contain a large number of genes as features after applying differential gene expression, and LASSO regression identified the most relevant genes for cancer type prediction whereas it reduced the features obtained using differential gene expression from 26,768 to 520. For DNA methylation data, the number of features (methylation sites) was reduced from 139,321 to 393. The pipeline for data processing is summarized in Table 2.

**Table 2.** Pipeline for data processing. Row 1 (Original Features): number of original features, Row 2 (Differentially Expressed Genes): number of features after applying Differentially Expressed Analysis, Row 3 (LIMMA model): number of features after applying LIMMA model, Row 4 (LASSO Regression Model): number of features after applying LASSO Regression Model, Row 5 (All Tumor Samples and Normal): number of tumor samples and normal samples, Row 6 (Unique Tumor Samples and Normal): number of tumor and normal samples after removing duplicates, Row 7 (Common Samples and Features): number of tumor and normal samples and features common across all datatypes, and Row 8 (Network Nodes and Edges): number of nodes and edges for the PPI network for each datatype.

| Datatype | mRNA | miRNA | DNA Methylation |
|---|---|---|---|
| Original Features | 60,660 | 1881 | 485,577 |
| Differentially Expressed Analysis | 26,768 | - | - |
| LIMMA Model (Selected Features) | - | - | 139,321 |
| LASSO Regression Model (Selected Features) | 520 | - | 393 |
| All Tumor Samples and Normal | 10,668 | 10,465 | 9,171 |
| Unique Tumor Samples and Normal | 10,667 | 10,465 | 8,674 |
| Common Samples and Features | 8,464 Samples and 2,794 Features | | |
| Network Nodes and Edges | 504 Nodes and 343 Edges | | |

## 3.3 Multi-Omics Data Integration

We integrated mRNA or RNA-Seq, miRNA, and DNA methylation data based on the sample ID, where the goal was to combine the datasets so that the omics data pertaining to each sample are merged into a single record. To achieve the integration, we utilized an inner join operation on the sample ID column across the three datasets. An inner join retains only the samples that have data in all three datasets. This approach ensures that the integrated dataset contains only samples for which complete omics data are available, thereby avoiding missing data issues. By merging the datasets using an inner join, we created a unified dataset that incorporates mRNA or RNA-Seq, miRNA, and DNA methylation data for each sample. During the integration process, we encountered instances where certain cancer types did not have data available for specific omics layers. For example, the RNA-Seq data was null for the cancer type TCGA_LAML, and the miRNA data was null for the cancer type TCGA_GBM. These cancer types were excluded from the integrated dataset to ensure that only samples with complete omics data were retained. Despite these exclusions, the integrated dataset still encompasses a diverse range of cancer types, totaling 31 cancer types along with normal samples. In total, the integrated dataset comprises 8,464 samples, each with 2,794 omics features. The preprocessing steps and data integration are depicted in Figure 1.

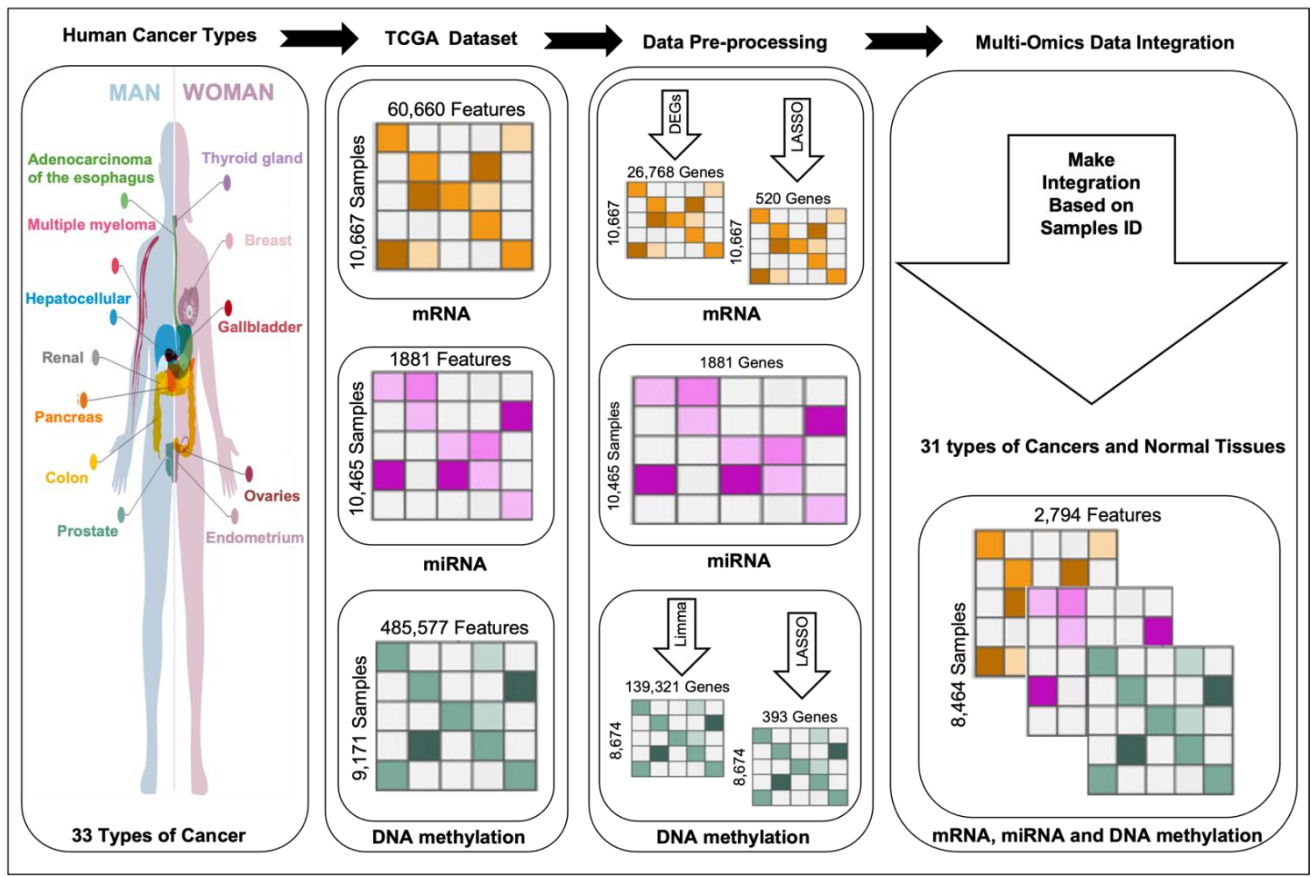

**Figure 1.** Preprocessing steps and data integration: First, omics data (mRNA, miRNA, and DNA methylation) were obtained from the Pan-cancer Atlas using the TCGAbiolinks library. Next, differential expression analysis (DEG) and LASSO regression were applied to mRNA data, while LIMMA and LASSO regression were applied to DNA methylation data. Subsequently, mRNA or RNA-Seq, miRNA, and DNA methylation data were integrated based on the sample ID using an inner join operation.

## 3.4 Graph Attention Network

Graph Attention Networks (GATs) employ attention mechanisms to enhance the functionality of standard GNNs [48]. GATs are particularly well-suited for handling data with complex relationships, such as multi-omics data, where features can be represented as nodes in a graph [49]. GATs function by computing attention coefficients for every edge in the graph, which enables the network to determine which nodes are more important to the representation of each node. Through the use of attention mechanism, GATs focus on pertinent nodes while simultaneously gathering information from nearby nodes. The propagation in a GAT layer is described by:

$$h_i^{(l+1)} = \sigma\left(\sum_{j \in \mathcal{N}(i)} \alpha_{ij}^{(l)} W^{(l)} h_j^{(l)}\right) \qquad (2)$$

where $h_i^{(l)}$ is the representation of node $i$ in layer $l$, $\mathcal{N}(i)$ is the neighborhood of node $i$, $\alpha_{ij}^{(l)}$ is the attention coefficient for edge $i \rightarrow j$ in layer $l$, $W^{(l)}$ is the weight matrix for layer $l$, and $\sigma$ is the activation function.

The attention coefficients $\alpha_{ij}^{(l)}$ are computed using a softmax function over all neighbors of node $i$ as follows:

$$\alpha_{ij}^{(l)} = \text{softmax}_j\left(\text{LeakyReLU}\left(\vec{a}^{(l)^T} [W^{(l)} h_i^{(l)} \| W^{(l)} h_j^{(l)}]\right)\right) \qquad (3)$$

where $\|$ denotes concatenation, $\vec{a}^{(l)}$ is a weight vector to be learned, and LeakyReLU is the activation function.

In the context of multi-omics data, the importance of GATs is due to their capacity to efficiently extract and combine data from several omics layers that are represented graphically. More accurate and transparent models in multi-omics data analysis are made possible by the ability to learn which characteristics or nodes are more pertinent for each sample's representation through the use of attention mechanisms.

We designed a GAT model to operate on PPI networks represented as graphs. The model takes as input the PPI network structure and node features, where each node represents a protein, and the edges represent interactions between proteins. The node features include multi-omics data such as gene expression, miRNA expression, and DNA methylation levels. The GAT architecture, depicted in Figure 2, consists of four GAT Convolutional layers with batch normalization and LeakyReLU activation functions (with a default negative slope parameter of 0.01) applied after each GAT Convolutional layer to normalize and activate the outputs. A dropout layer with a rate of 0.5 is added after each GAT Convolutional layer to reduce overfitting by randomly dropping a fraction of the units during training. In the forward pass, the input data is passed through each GAT Convolutional layer, followed by batch normalization, activation, and dropout. The final output is passed through a softmax function to obtain class probabilities. Our model was trained and evaluated using 5-fold cross-validation to ensure robust performance metrics and avoid overfitting. For each fold, the model was initialized and trained from scratch, with the parameters optimized using the training set and performance evaluated on the corresponding test set. The dataset was split into five subsets, with each subset used once as a test set while the remaining four subsets were used for training. This process was repeated five times, and the average performance metrics across all folds were reported. The loss function used for our method, as described in our model implementation, is the Negative Log-Likelihood Loss. This loss function is suitable for multi-class classification tasks and is applied to the output of the final fully connected layer of our model, which utilizes the Graph Attention Network (GAT) layers for feature aggregation and classification.

We designed a GAT model to operate on PPI networks represented as graphs. The model takes as input the PPI network structure and node features, where each node represents a protein, and the edges represent interactions between proteins. The node features include the multi-omics data gene expression, miRNA expression, and DNA methylation levels. The GAT architecture is depicted in Figure 2 and consists of four GAT Convolutional layers with batch normalization and LeakyReLU activation functions applied after each GAT Convolutional layer to normalize and activate the outputs. A dropout layer is added to reduce overfitting by randomly dropping a fraction of the units during training. I.e., in the forward pass, the input data is passed through each GAT Convolutional layer, followed by batch normalization, activation, and dropout. The final output is passed through a softmax function to obtain class probabilities.

### 3.5 Protein-Protein Interaction (PPI) Network

Protein-Protein Interaction (PPI) network is a graphical representation of the connections between proteins in a cell [50]. Proteins work together to perform a variety of biological tasks, including metabolism, gene control, and signaling [51]. Understanding the molecular mechanisms behind cellular functions and illnesses, such as cancer, is made possible with the use of PPI networks [52]. A valuable tool for obtaining anticipated and experimentally verified PPI is the String database [53]. For the used TCGA cancer types, the PPI network is shown in Figure 3. To create the PPI network, we retrieved the protein interaction data for the genes of the mRNA or RNA-Seq data from the String database. The data include information regarding the interacting proteins and the confidence scores of the interactions. Next, the data is processed to build the PPI network as a graph, where each protein is represented as a node and the interactions between proteins are represented as edges. Subsequently, the PPI network is inputted as a graph structure for GAT by coding the edges and nodes as tensors. Each patient is represented by the topology of a PPI network in the graph structure. Multi-Omics features from mRNA, miRNA, and DNA methylation data enrich the nodes. Although every patient's network has the same topology, the node feature values differ, representing their unique cancer-specific molecular patterns.

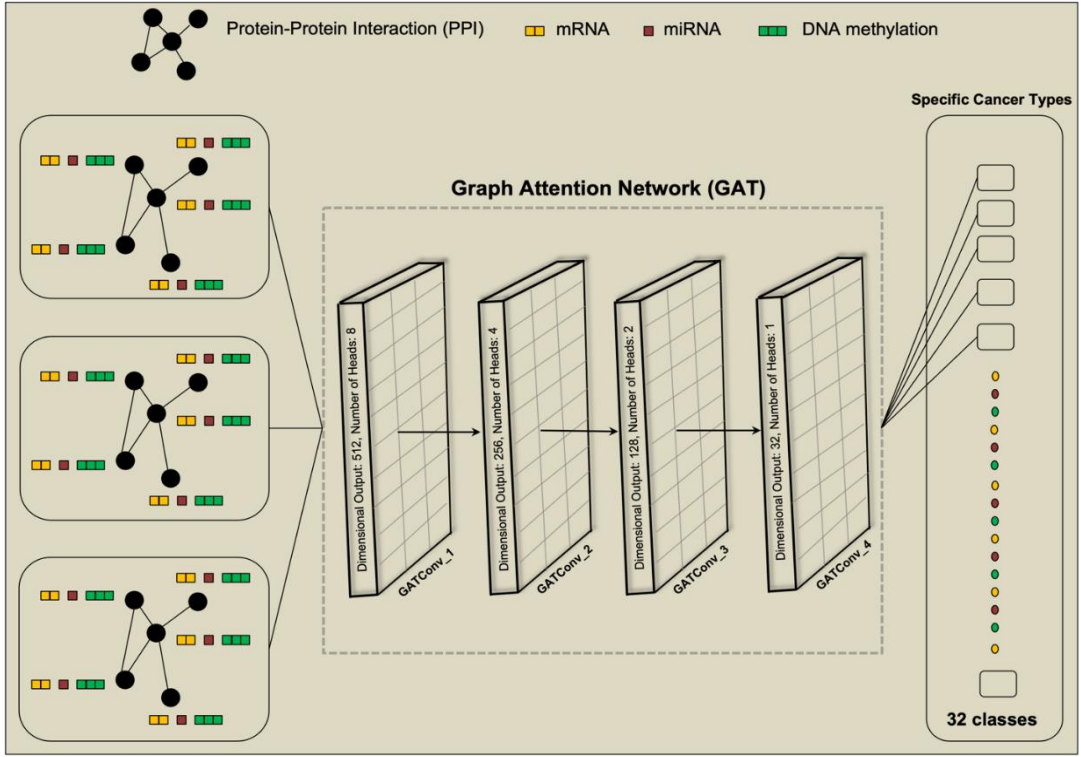

**Figure 2.** The architecture of the GAT model for multiclass cancer classification.

### 3.6 Evaluation metrics

For model evaluation, we used accuracy, precision, recall, and F1 score performance metrics. Accuracy is computed using the following equation and pertains to the corrected classified cancer type:

$$\text{Accuracy} = \frac{TP+TN}{TP+TN+FP+FN} \times 100\% \qquad (4)$$

True Positives (TP) denote the proportion of samples correctly classified as positive, True Negatives (TN) denote the proportion of samples correctly predicted as negative, False Positives (FP) denote the proportion of negative samples incorrectly predicted as positive, and False Negatives (FN) represent the proportion of positive samples incorrectly predicted as negative.

Precision measures the ratio of true positives to total correct classification, where high precision indicates that the model performs well in classifying true positives. Recall is the ratio of true positives to all positives, where high recall indicates that the model can effectively discriminate between cancer types. Precision and recall have trade-offs, therefore improving only one metric does not necessarily result in a more efficient model. F1 score is the harmonic mean of recall and precision, as represents both recall and precision in one metric. The metrics are frequently expanded to handle a multitude of classes in multi-class classification scenarios through the use of macro-averaging (MA) or micro-averaging techniques, which offer a means of summarizing the performance across numerous classes. While micro-averaging takes into account the overall counts of true positives, false positives, and false negatives across all classes, MA treats each class equally. We selected MA, which computes metrics independently for each class and then averages them, because it gives each class's performance equal weight. Since MA is insensitive to class imbalance and fairly represents overall model performance, every class contributes equally to the final average. MA for recall, precision, and F1 score are calculated as follows:

$$\text{Recall } (MA) = \frac{1}{K} \sum_{i=1}^{k} \frac{TP_i}{TP_i + FN_i} \times 100\% \qquad (5)$$

$$\text{Precision } (MA) = \frac{1}{K} \sum_{i=1}^{k} \frac{TP_i}{TP_i + FP_i} \times 100\% \qquad (6)$$

$$F1 \ (MA) = \frac{1}{K} \sum_{i=1}^{k} 2 \times \frac{\text{Recall}_i \times \text{Precision}_i}{\text{Recall}_i + \text{Precision}_i} \times 100\% \qquad (7)$$

where $k$ denotes the total number of classes. In our case $k = 32$ including the normal samples.

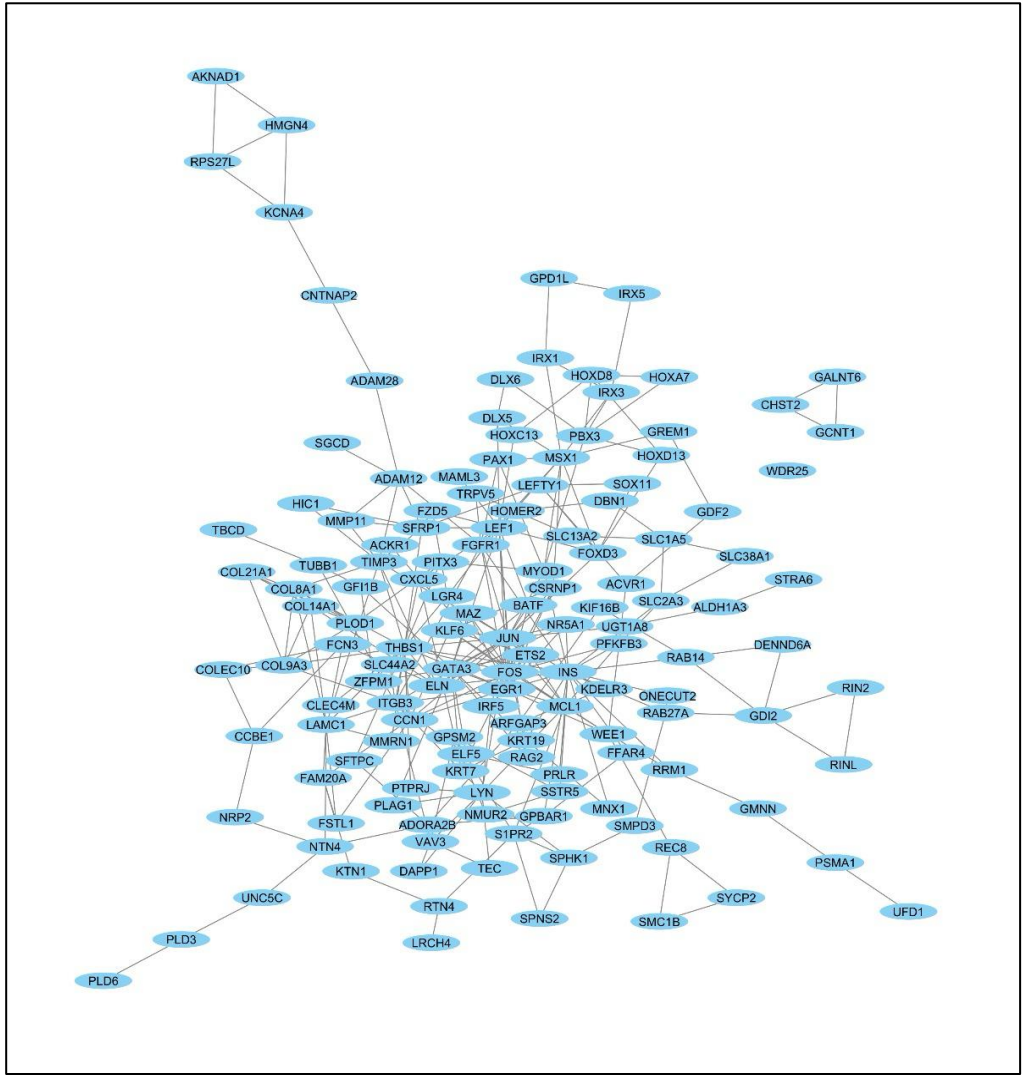

**Figure 3.** Protein-Protein Interaction (PPI) Network.

## 4. Experimental Results

The values of the performance metrics obtained with the proposed LASSO-MOGAT approach based on PPI networks for cancer classification are presented in Table 3. The integration of multi-omics data, including mRNA/RNA-Seq, miRNA, and DNA methylation, yields the highest accuracy of 94.68%, along with precision of 90.38%, recall of 89.87%, and F1 score of 89.87%. The results imply that a representation of cancer biology through multi-omics integration enhances classification accuracy and maintains a well-balanced model with high precision and recall, crucial for medical applications. This aligns with the systems biology concept, emphasizing the importance of considering interactions between different biological components.

The results in Table 3 indicate that models based on single omics data types, while still effective, exhibit lower accuracies in comparison to multi-omics models. For example, the model with miRNA data performs the poorest with an accuracy of 87.61%, precision of 81.17%, recall of 81.88%, and F1 score of 81.26%. This underlines the advantage of multi-omics integration in improving classification models, as seen in the higher precision, recall, and F1 score values for the multi-omics models.

Interestingly, certain combinations of omics data types, such as mRNA/RNA-Seq with miRNA or DNA methylation, do not provide significant performance improvements, highlighting the complexity of omics data interactions. Conversely, the combination of mRNA and DNA methylation in multi-omics data achieves a high accuracy of 94.05%, with precision of 89.66%, recall of 89.54%, and F1 score of 89.35%, indicating potential synergistic effects between these data types.

Overall, the findings suggest that while single omics data types can be effective, integrating multiple omics data types, particularly with GATs based on PPI networks, offers a more comprehensive and accurate approach to cancer classification.

**Table 3.** Performance metrics of the proposed LASSO-MOGAT approach based on PPI Network.

| Data Types | Multi-Omics Data Type | Cancer Types | Accuracy Mean ± std | Precision | Recall | F1 Score |
|---|---|---|---|---|---|---|
| Single Omics Data | mRNA or RNA-Seq | 31 Types of Cancer Plus Normal Tissues | 92.08% ± 0.0043 | 0.8769 | 0.8716 | 0.8674 |
| Single Omics Data | miRNA | 31 Types of Cancer Plus Normal Tissues | 87.61% ± 0.0083 | 0.8117 | 0.8188 | 0.8126 |
| Single Omics Data | DNA methylation | 31 Types of Cancer Plus Normal Tissues | 92.23% ± 0.0094 | 0.8770 | 0.8599 | 0.8580 |
| Multi-Omics Data | mRNA or RNA-Seq and miRNA | 31 Types of Cancer Plus Normal Tissues | 92.78% ± 0.0065 | 0.8902 | 0.8902 | 0.8882 |
| Multi-Omics Data | mRNA or RNA-Seq and DNA methylation | 31 Types of Cancer Plus Normal Tissues | 94.05% ± 0.0039 | 0.8966 | 0.8954 | 0.8935 |
| Multi-Omics Data | miRNA and DNA methylation | 31 Types of Cancer Plus Normal Tissues | 93.82% ± 0.0046 | 0.9035 | 0.8926 | 0.8926 |
| **Multi-Omics Data** | mRNA or RNA-Seq miRNA DNA methylation | 31 Types of Cancer Plus Normal Tissues | 94.68% ± 0.0060 | 0.9038 | 0.8987 | 0.8987 |

Table 4 shows the comparison between our proposed LASSO-MOGAT model and related models for classification of different types of cancer. The models by Mostavi et al. [22] based on 1D-CNN, 2D-Vanilla-CNN, and 2D-Hybrid-CNN exhibit high performance using mRNA data for 33 cancer types, with accuracies of 95.50%, 94.87%, and 95.70% respectively, demonstrating the capability of CNNs in processing omics data. The graph-based model with Transformer architecture GTN by Kaczmarek et al. [24], combines mRNA and miRNA data for 12 cancer types and achieves an accuracy of 93.56%. Similarly, GCNN by Ramirez et al. [23] based on PPI networks achieves accuracies of 89.99% and 94.71% respectively. These models highlight the significance of adding graph structures and multi-omics data to improve classification accuracy. By leveraging a large dataset comprising DNA methylation, miRNA, and mRNA levels for 31 different forms of cancer as well as normal tissues, the proposed LASSO-MOGAT model achieves an accuracy of 94.68% in cancer classification.

**Table 4.** Performance metrics for related multi-omics graph methods based on PPI Network.

| Authors | Models | Pan-Cancer | Multi-Omics Data type | | | Accuracy Mean ± std | Precision | Recall | F1 Score |
|---|---|---|---|---|---|---|---|---|---|
| | | | mRNA | miRNA | DNA methylation | | | | |
| Mostavi et al., 2020 [22] | 1D-CNN | 33 Types of Cancer and Normal Tissues | √ | | | 95.50% ± 0.100 | - | - | - |
| | 2D-Vanilla-CNN | | | | | 94.87 % ± 0.040 | - | - | - |
| | 2D-Hybrid-CNN | | | | | 95.70% ± 0.100 | - | - | - |

| | | | | | | | | |
|---|---|---|---|---|---|---|---|---|
| Ramirez et al., 2020 [23] | GCNN-PPI graph | 33 Types of Cancer and normal Tissues | √ | | | 89.99% ± 0.883 | 87.75% | 83.79% | - |
| | GCNN-PPI + singleton graph | | | | | 94.71% ± 0.107 | 92.76% | 92.19% | - |
| Kaczmarek et al., 2022 [24] | GTN | 12 Types of Cancer | √ | √ | | 93.56% ± 0.910 | - | - | - |
| **Proposed LASSO -MOGAT** | | 31 Types of Cancers and Normal Tissues | √ | √ | √ | 94.68% ± 0.006 | 90.38% | 89.87% | 89.87% |

## 5. Conclusion

This study demonstrates the potential of integrating multi-omics data with advanced graph-based deep learning models for cancer classification. The proposed LASSO-MOGAT framework effectively combines RNA-Seq, miRNA, and DNA methylation data with Graph Attention Networks and protein-protein interaction (PPI) networks to capture the intricate relationships within cancer biology. The experimental results validated through five-fold cross-validation highlight the performance of LASSO-MOGAT in terms of accuracy, precision, recall, and F1 score. By employing differential expression analysis with LIMMA and LASSO regression for feature selection, this method offers a robust and comprehensive understanding of cancer molecular mechanisms. These findings underscore the critical role of multi-omics integration and graph-based models in enhancing cancer classification accuracy and reliability, paving the way for improved diagnostic and therapeutic strategies in precision oncology.

## References:

1. Alharbi, F., & Vakanski, A. (2023). Machine learning methods for cancer classification using gene expression data: a review. Bioengineering, 10(2), 173.
2. Pfeifer, B., Saranti, A., & Holzinger, A. (2022). GNN-SubNet: disease subnetwork detection with explainable graph neural networks. Bioinformatics, 38(Supplement_2), ii120-ii126.
3. Wekesa, J. S., & Kimwele, M. (2023). A review of multi-omics data integration through deep learning approaches for disease diagnosis, prognosis, and treatment. Frontiers in Genetics, 14, 1199087.
4. Leng, D., Zheng, L., Wen, Y., Zhang, Y., Wu, L., Wang, J., ... & Bo, X. (2022). A benchmark study of deep learning-based multi-omics data fusion methods for cancer. Genome biology, 23(1), 171.
5. Gogoshin, G., & Rodin, A. S. (2023). Graph Neural Networks in Cancer and Oncology Research: Emerging and Future Trends. Cancers, 15(24), 5858.
6. Li, B., & Nabavi, S. (2024). A multimodal graph neural network framework for cancer molecular subtype classification. BMC bioinformatics, 25(1), 27.
7. Tanvir, R. B., Islam, M. M., Sobhan, M., Luo, D., & Mondal, A. M. (2024). MOGAT: A Multi-Omics Integration Framework Using Graph Attention Networks for Cancer Subtype Prediction. International Journal of Molecular Sciences, 25(5), 2788.
8. Narrandes, S., & Xu, W. (2018). Gene expression detection assay for cancer clinical use. Journal of Cancer, 9(13), 2249.
9. Singh, K. P., Miaskowski, C., Dhruva, A. A., Flowers, E., & Kober, K. M. (2018). Mechanisms and measurement of changes in gene expression. Biological research for nursing, 20(4), 369-382.
10. Li, M., Sun, Q., & Wang, X. (2017). Transcriptional landscape of human cancers. Oncotarget, 8(21), 34534.
11. Heo, Y. J., Hwa, C., Lee, G. H., Park, J. M., & An, J. Y. (2021). Integrative multi-omics approaches in cancer research: from biological networks to clinical subtypes. Molecules and cells, 44(7), 433-443.
12. Menyhárt, O., & Győrffy, B. (2021). Multi-omics approaches in cancer research with applications in tumor subtyping, prognosis, and diagnosis. Computational and structural biotechnology journal, 19, 949-960.
13. Geissler, F., Nesic, K., Kondrashova, O., Dobrovic, A., Swisher, E. M., Scott, C. L., & J. Wakefield, M. (2024). The role of aberrant DNA methylation in cancer initiation and clinical impacts. Therapeutic Advances in Medical Oncology, 16, 17588359231220511.
14. Ankasha, S. J., Shafiee, M. N., Wahab, N. A., Ali, R. A. R., & Mokhtar, N. M. (2018). Post-transcriptional regulation of microRNAs in cancer: From prediction to validation. Oncology reviews, 12(1).

15. Mohamed, T. I., & Ezugwu, A. E. (2024). Enhancing Lung Cancer Classification and Prediction with Deep Learning and Multi-Omics Data. IEEE Access.
16. Qiu, L., Li, H., Wang, M., & Wang, X. (2021). Gated Graph Attention Network for Cancer Prediction. Sensors, 21(6), 1938.
17. Baul, S., Ahmed, K. T., Filipek, J., & Zhang, W. (2022). omicsGAT: Graph attention network for cancer subtype analyses. International Journal of Molecular Sciences, 23(18), 10220.
18. Zhao, W., Gu, X., Chen, S., Wu, J., & Zhou, Z. (2022). MODIG: integrating multi-omics and multi-dimensional gene network for cancer driver gene identification based on graph attention network model. Bioinformatics, 38(21), 4901-4907.
19. Jeong, D., Koo, B., Oh, M., Kim, T. B., & Kim, S. (2023). GOAT: Gene-level biomarker discovery from multi-Omics data using graph ATtention neural network for eosinophilic asthma subtype. Bioinformatics, 39(10), btad582.
20. Shi, H., Gu, Y., Zhang, H., Li, X., & Cao, Y. (2023, July). MORGAT: A Model Based Knowledge-Informed Multi-omics Integration and Robust Graph Attention Network for Molecular Subtyping of Cancer. In International Conference on Intelligent Computing (pp. 192-206). Singapore: Springer Nature Singapore.
21. Yang, K., Cheng, J., Cao, S., Pan, X., Shen, H. B., Jin, C., & Yuan, Y. (2023). Integration of multi-source gene interaction networks and omics data with graph attention networks to identify novel disease genes. bioRxiv, 2023-12.
22. Mostavi, M., Chiu, Y. C., Huang, Y., & Chen, Y. (2020). Convolutional neural network models for cancer type prediction based on gene expression. BMC medical genomics, 13, 1-13.
23. Ramirez, R., Chiu, Y. C., Hererra, A., Mostavi, M., Ramirez, J., Chen, Y., ... & Jin, Y. F. (2020). Classification of cancer types using graph convolutional neural networks. Frontiers in physics, 8, 203.
24. Kaczmarek, E., Jamzad, A., Imtiaz, T., Nanayakkara, J., Renwick, N., & Mousavi, P. (2021). Multi-omic graph transformers for cancer classification and interpretation. In PACIFIC SYMPOSIUM ON BIOCOMPUTING 2022 (pp. 373-384).
25. Moon, S., & Lee, H. (2022). MOMA: a multi-task attention learning algorithm for multi-omics data interpretation and classification. Bioinformatics, 38(8), 2287-2296.
26. Zhang, G., Peng, Z., Yan, C., Wang, J., Luo, J., & Luo, H. (2022). MultiGATAE: a novel cancer subtype identification method based on multi-omics and attention mechanism. Frontiers in Genetics, 13, 855629.
27. Sun, Q., Cheng, L., & Zhang, L. (2023). SADLN: Self-attention based deep learning network of integrating multi-omics data for cancer subtype recognition. Frontiers in Genetics, 13, 1032768.
28. Shanthamallu, U. S., Thiagarajan, J. J., Song, H., & Spanias, A. (2019). Gramme: Semisupervised learning using multilayered graph attention models. IEEE transactions on neural networks and learning systems, 31(10), 3977-3988.
29. Ouyang, D., Liang, Y., Li, L., Ai, N., Lu, S., Yu, M., ... & Xie, S. (2023). Integration of multi-omics data using adaptive graph learning and attention mechanism for patient classification and biomarker identification. Computers in Biology and Medicine, 164, 107303.
30. Gong, P., Cheng, L., Zhang, Z., Meng, A., Li, E., Chen, J., & Zhang, L. (2023). Multi-omics integration method based on attention deep learning network for biomedical data classification. Computer Methods and Programs in Biomedicine, 231, 107377.
31. Song, H., Yin, C., Li, Z., Feng, K., Cao, Y., Gu, Y., & Sun, H. (2023). Identification of Cancer Driver Genes by Integrating Multiomics Data with Graph Neural Networks. Metabolites, 13(3), 339.
32. Zhang, X., Zhang, J., Sun, K., Yang, X., Dai, C., & Guo, Y. (2019, November). Integrated multi-omics analysis using variational autoencoders: application to pan-cancer classification. In 2019 IEEE International Conference on Bioinformatics and Biomedicine (BIBM) (pp. 765-769). IEEE.
33. Chai, H., Zhou, X., Zhang, Z., Rao, J., Zhao, H., & Yang, Y. (2021). Integrating multi-omics data through deep learning for accurate cancer prognosis prediction. Computers in biology and medicine, 134, 104481.
34. Li, X., Ma, J., & Leng, L. MoGCN: a multi-omics integration method based on graph convolutional network for cancer subtype analysis. Front Genet. 2022; 13: 806842.
35. Zhou, N., Wang, S., & Tan, Z. (2022, October). AEMVC: anchor enhanced multi-omics cancer subtype identification. In Proceedings of the 3rd International Symposium on Artificial Intelligence for Medicine Sciences (pp. 57-63).
36. Khadirnaikar, S., Shukla, S., & Prasanna, S. R. M. (2023). Integration of pan-cancer multi-omics data for novel mixed subgroup identification using machine learning methods. Plos one, 18(10), e0287176.
37. Zhu, J., Oh, J. H., Simhal, A. K., Elkin, R., Norton, L., Deasy, J. O., & Tannenbaum, A. (2023). Geometric graph neural networks on multi-omics data to predict cancer survival outcomes. Computers in biology and medicine, 163, 107117.
38. Xiao, S., Lin, H., Wang, C., Wang, S., & Rajapakse, J. C. (2023). Graph neural networks with multiple prior knowledge for multi-omics data analysis. IEEE Journal of Biomedical and Health Informatics.
39. Chatzianastasis, M., Vazirgiannis, M., & Zhang, Z. (2023). Explainable multilayer graph neural network for cancer gene prediction. Bioinformatics, 39(11), btad643.
40. Wang, J., Liao, N., Du, X., Chen, Q., & Wei, B. (2024). A semi-supervised approach for the integration of multi-omics data based on transformer multi-head self-attention mechanism and graph convolutional networks. BMC genomics, 25(1), 86.

41. Yao, D., Li, B., Zhan, X., Zhan, X., & Yu, L. (2024). GCNFORMER: graph convolutional network and transformer for predicting lncRNA-disease associations. BMC bioinformatics, 25(1), 5.
42. Weinstein, J. N., Collisson, E. A., Mills, G. B., Shaw, K. R., Ozenberger, B. A., Ellrott, K., ... & Stuart, J. M. (2013). The cancer genome atlas pan-cancer analysis project. Nature genetics, 45(10), 1113-1120.
43. Colaprico, A., Silva, T. C., Olsen, C., Garofano, L., Cava, C., Garolini, D., ... & Noushmehr, H. (2016). TCGAbiolinks: an R/Bioconductor package for integrative analysis of TCGA data. Nucleic acids research, 44(8), e71-e71.
44. Rapaport, F., Khanin, R., Liang, Y., Pirun, M., Krek, A., Zumbo, P., ... & Betel, D. (2013). Comprehensive evaluation of differential gene expression analysis methods for RNA-seq data. Genome biology, 14, 1-13.
45. Chen, J., Long, M. D., Sribenja, S., Ma, S. J., Yan, L., Hu, Q., ... & Yao, S. (2023). An epigenome-wide analysis of socioeconomic position and tumor DNA methylation in breast cancer patients. Clinical Epigenetics, 15(1), 68.
46. Pidsley, R., Y Wong, C. C., Volta, M., Lunnon, K., Mill, J., & Schalkwyk, L. C. (2013). A data-driven approach to preprocessing Illumina 450K methylation array data. BMC genomics, 14, 1-10.
47. Wang, Z., Fu, G., Ma, G., Wang, C., Wang, Q., Lu, C., ... & Li, S. (2024). The association between DNA methylation and human height and a prospective model of DNA methylation-based height prediction. Human Genetics, 1-21.
48. Sheng, J.; Zhang, Y.; Wang, B.; Chang, Y. MGATs: Motif-Based Graph Attention Networks. Mathematics 2024, 12, 293. https://doi.org/10.3390/math12020293
49. Lazaros, K., Koumadorakis, D.E., Vlamos, P. et al. Graph neural network approaches for single-cell data: a recent overview. Neural Comput & Applic (2024). https://doi.org/10.1007/s00521-024-09662-6
50. Zainal-Abidin RA, Afiqah-Aleng N, Abdullah-Zawawi MR, Harun S, Mohamed-Hussein ZA. Protein-Protein Interaction (PPI) Network of Zebrafish Oestrogen Receptors: A Bioinformatics Workflow. Life (Basel). 2022 Apr 27;12(5):650. doi: 10.3390/life12050650. PMID: 35629318; PMCID: PMC9143887.
51. Morris R, Black KA, Stollar EJ. Uncovering protein function: from classification to complexes. Essays Biochem. 2022 Aug 10;66(3):255-285. doi: 10.1042/EBC20200108. PMID: 35946411; PMCID: PMC9400073.
52. Hu, H., Wang, H., Yang, X. et al. Network pharmacology analysis reveals potential targets and mechanisms of proton pump inhibitors in breast cancer with diabetes. Sci Rep 13, 7623 (2023). https://doi.org/10.1038/s41598-023-34524-x.
53. Damian Szklarczyk, Annika L Gable, Katerina C Nastou, David Lyon, Rebecca Kirsch, Sampo Pyysalo, Nadezhda T Doncheva, Marc Legeay, Tao Fang, Peer Bork, Lars J Jensen, Christian von Mering, The STRING database in 2021: customizable protein–protein networks, and functional characterization of user-uploaded gene/measurement sets, Nucleic Acids Research, Volume 49, Issue D1, 8 January 2021, Pages D605–D612.